\documentclass[10pt, journal]{IEEEtran}
\usepackage{blindtext, graphicx}
\usepackage{algorithm}
\usepackage{algorithmic}
\usepackage{bbding}
\usepackage{amsmath}
\usepackage{bbold}
\usepackage{hyperref}
\usepackage{amsfonts,amssymb}
\usepackage{cite}
\usepackage{float}

\newcommand{\argmin}{\operatornamewithlimits{argmin}}

\graphicspath{{figures/}}

\begin{document}

\title{Learning Constructive Primitives for Online Level Generation and Real-time Content Adaptation in Super Mario Bros}

\author{Peizhi~Shi
        and~Ke~Chen,~\IEEEmembership{Senior~Member,~IEEE}

\thanks{The authors are with School of Computer Science, The University of Manchester, Manchester M13 9PL, United Kingdom (e-mail: shipa@cs.manchester.ac.uk; chen@cs.manchester.ac.uk).}}

\maketitle

\begin{abstract}
 Procedural content generation (PCG) is of great interest to game design and development as it generates game content automatically.
 Motivated by the recent learning-based PCG framework and other existing PCG works, we propose an alternative approach to online content generation and adaptation in \emph{Super Mario Bros} (SMB). Unlike most of existing works in SMB, our approach exploits the synergy between rule-based and learning-based methods to produce constructive primitives, quality yet controllable game segments in SMB. As a result, a complete quality game level can be generated online by integrating relevant constructive primitives via controllable parameters regarding geometrical features and procedure-level properties. Also the adaptive content can be generated in real time by dynamically selecting proper constructive primitives via an adaptation criterion, e.g., dynamic difficulty adjustment (DDA). Our approach is of several favorable properties in terms of content quality assurance, generation efficiency and controllability. Extensive simulation results demonstrate that the proposed approach can generate controllable yet quality game levels online and adaptable content for DDA in real time.
\end{abstract}


\IEEEpeerreviewmaketitle

\setlength{\baselineskip}{12pt}
\section{Introduction}
\par
\IEEEPARstart{I}{n} video gaming industry, game content construction and generation are laborious and costed enormously. As a potential solution, procedural content generation (PCG) \cite{SBPCG} is a game design and development methodology that aims generating game content automatically via algorithms, which receives a lot of attention from academic research communities and game industries. It is anticipated that the proper use of PCG techniques would reduce the cost of game design and development dramatically. Moreover, PCG could also provide a way that automatically generates personalized games that can adapt content towards a player's preference and optimizing their cognitive/affective experience.
\par
In the PCG research, a number of methods have been proposed for game content generation. These techniques have been applied to different game genres ranging from platform games to first person shooter (see \cite{SBPCG} for review) with different styles, e.g., online or offline content generation \cite{SBPCG}. Online style tends to speed up content construction process, while offline style can be used for endless or adaptive game generation. Although wide varieties of PCG techniques have been proposed, there are still some open challenges for content generation, including game quality assurance \cite{PCGGCAS}, generation efficiency \cite{SBPCG}, and controllability \cite{PCGGCAS}.
\par
\textit{Super Mario Bros} (SMB), a classic 2D platform game, has become a popular test bed for PCG-related researches \cite{MARIO}. In this platform game, a player runs from the left side of the screen to the right side, fights enemies, and rescues the \emph{Princess Peach}. SMB has a number of different game elements (e.g., enemies, coins, tubes and cannons) that can be generated via PCG. In recent years, several SMB level generators have been developed for \emph{level generation track of the Mario AI Championship} \cite{MAIC} as well as presented in publications \cite{LAUNCHPAD, AGACMPCVGL, ELSMBGE, AMLLG, PLGORE, EMGMC}. While those level generators can generate complete levels in SMB, there are still several issues to be addressed:

\begin{itemize}
\item \emph{Quality assurance}. The quality of procedural SMB levels is generally not as high as handcrafted levels in real games \cite{PCGGCAS}; levels generated by existing generators often contain aesthetically unappealing items (e.g., weird enemy/coin/box decoration), unplayable or unbalanced structures, unreachable resources (e.g., coins and boxes), and unexplainable difficulty curves. Such problems could result in negative gameplay experience.
\item \emph{Efficiency}. The generation process of some SMB generators (e.g., \cite{PLGORE, AGACMPCVGL}) is computationally expensive; it may take several seconds to generate a complete game level. Thus, such generators are not suitable for online generation.
\item \emph{Controllability}. The geometrical features (e.g., coordinate of each enemy and tube) and some properties of procedural levels (e.g., linearity \cite{AERLG} and density \cite{ELSMBGE}) are not directly controllable in a number of existing level generators. Instead a game designer has to encode the desired properties in handcrafted rules (e.g., \cite{LAUNCHPAD}) in order to evaluate or control the procedural levels \cite{PCGGCAS}. However, those heuristic rules may be difficult to formulate \cite{EDPCG} and could even slow down the level generation \cite{ACEPLGM}.
\end{itemize}
\par
Unlike most of the existing works, we propose an alternative approach in this paper in order to tackle the aforementioned issues. Motivated by the learning-based PCG (LBPCG) framework \cite{LBPCG} and other existing works \cite{LAUNCHPAD, AGACMPCVGL, ELSMBGE, AMLLG, PLGORE, EMGMC}, we explore the content space in SMB from a different perspective by taking short game segments into account. To address the quality assurance issue, we exploit the synergy between rule-based and learning-based methods; easy-to-design rules are employed for removal of apparently unappealing game segments, and then active learning by encoding a game designer's knowledge implicitly is applied to obtain high quality game segments, hereinafter named \emph{constructive primitives} (CPs). Those CPs  not only provide quality building blocks  but also enable to control the local geometry and the level properties effectively in the SMB level generation. As a result, a complete quality game level can be efficiently generated online by integrating relevant constructive primitive together via controllable parameters on geometrical features, e.g., coordinates of enemies and cannons, and level properties, e.g., linearity, density and leniency. Moreover, adaptable content can be generated in real time by choosing the proper CPs with a given adaptation criterion, e.g., dynamic difficulty adjustment by matching content difficulty and player's performance.  It is worth stating that our approach presented in this paper significantly distinguishes from existing segment-based works in SMB (e.g., \cite{LAUNCHPAD}) in terms of quality assurance, efficiency and controllability, which is discussed later on. Thus, learning CPs paves an alternative way in addressing those PCG issues in SMB.
\par
The main contributions of this paper are summarized as follows: a) a novel approach to producing quality yet controllable game segments or CPs in SMB; b) a controllable online level generator and enabling techniques for real-time content adaptation based on CPs; and c) a thorough evaluation on our proposed approach.
\par
The rest of paper is organized as follows. Sect. II reviews the previous works. Sect. III presents our approach to learning CPs in SMB. Sect. IV describes two methods for applications of CPs in online level generation and real-time content adaptation. Sect. V reports simulation results, and the last section discusses relevant issues and implications of this research.

\section{Related Work}
We review the relevant works in PCG and content adaptation that motivate our work presented in this paper.

\subsection{Procedural Content Generation}
\par
First of all, many SMB level generators were developed under the \emph{search-based procedural content generation} (SBPCG) framework \cite{SBPCG}. In such approaches, game developers first construct a content space that contains all possible procedural levels and then employ heuristic/stochastic search algorithms, e.g., genetic algorithm, to find out high quality levels from the content space. For instance, Sorenson et al. \cite{AGACMPCVGL} proposed a search-based SMB level generator for \emph{level generation track of the Mario AI Championship} \cite{MAIC}. Their approach used a set of handcrafted constraints and a challenge-based metric as evaluation functions. During content generation, game levels were first tested by these constraints. Survivals from the test would be used in the search-based optimization. As a result, only game levels of the highest challenge-related fitness values were released to players. In this approach, the properties of procedural levels can be controlled by parameters used in evaluation functions. However, their level generator has to take several minutes to generate 50 levels together due to the huge content space and the use of multiple evaluation functions that may slow down the level generation \cite{ACEPLGM}. In addition, identifying proper constraints and formulating heuristic evaluation functions need game developers' wisdom and great efforts. This process is often tedious and time-consuming as game content is observable but hard to explain and abstract. Hence, it is difficult to build up an explicit relationship between game content and its quality measurement via handcrafted constraints and heuristic evaluation functions. Similarly, other search-based SMB level generators, e.g., \cite{AGACMPCVGL, ELSMBGE, AMLLG}, suffer from the same problems and some of them do not adequately address the quality assurance issue pointed out in Sect. I.
\par
As an alternative methodology in PCG, \emph{constructive PCG} is also applied in the development of SMB level generators. In such approaches, a set of constructive rules are designed by human experts and then used to convert high-level game parameters, e.g., the number of enemies, game difficulty, style, and random seed to concrete game levels. Furthermore, constructive rules have to assure the game quality. \emph{Parameterized Notch} generator \cite{FAMGCQ} is a typical constructive level generator for SMB. In this approach, a procedural level is controlled and generated via constructive rules working on several content features, e.g., the number of enemies, the averaging width of gaps, and the spatial diversity of gaps. In comparison to the SBPCG approaches, creating constructive rules might be even more difficult \cite{EDPCG}; constructive rules are often more complicated than observable constraints. Hence, imperfect constructive rules may lead to low quality and limited controllability over game levels. This fundamental limitation also exists in other constructive SMB generators, e.g., \cite{NOTCH, MPESMB, MAIC, EMGMC}.
\par
Unlike the SBPCG and the constructive PCG approaches that generate a game level at a global level, there are SMB level generators that first generate segments by exploring the local properties and then merge segments to generate a complete game level. We name such level generators \emph{segment-based} approaches. For example, Smith et al. developed the \emph{Launchpad} level generator \cite{LAUNCHPAD} for generic 2D platform games including SMB. Instead of generating a complete level directly, they first used a grammar-based method that generates game segments named ``rhythm groups". Then they merged these rhythm groups together to form a complete level. Before releasing a generated level to player, they used several critics to examine whether the generated game level is acceptable. As a salient characteristic, this approach provides human designers/users with a variety of parameters to control the procedural levels ranging from frequencies of geometry components to level path equation. However, some procedural level properties, e.g., linearity and component frequencies, had to be controlled by designer-specified evaluation functions, which often slows down its generation process. In addition, coordinates of game elements are not controllable directly in this approach. Although our proposed approach is motivated by the segmented-based approaches, ours considerably distinguishes from the existing works in several aspects including the controllability of geometrical features (e.g., coordinate of each enemies, cannons and hills) and procedural level properties (e.g., density \cite{ELSMBGE}, leniency \cite{AERLG} and linearity \cite{AERLG}). In particular, our approach generates quality segments of the fixed-size via combining rule-based and learning-based evaluation functions rather than the handcrafted ones used in the existing segment-based methods, e.g., \cite{LAUNCHPAD}.

\subsection{Content Adaptation}

\par
Content adaptation is demanded in generating the personalized content to optimize cognitive/affective experience during gameplay \cite{EDPCG}. In general, such techniques can be applied to different types of game content ranging from non-player character behaviour or game AI to game geometry (see \cite{EDPCG} for a review). As our work concerns only the game geometry adaptation, we briefly review the relevant work regarding SMB in this context.

\par
A typical example of geometry adaptation is the \emph{personalized level generator} in SMB \cite{MPESMB, TAPCGPG}. This level generator is controllable so that various game levels can be generated with a number of controllable features. For content adaptation, an affective model was created to
map a player's behaviour and the controllable content features onto each of the player's affective states. With such a model, the level generator can generate personalized game levels for an individual. For an effective mapping, the controllable content features were selected based on their impact on the affective states. With the selected features, the game developers then designed a number of constructive rules to control these content features. As argued in Sect. II-A, proper constructive rules are difficult to be handcrafted and the use of evaluation functions to control a complete procedural level could slow down content generation. Similar to this work \cite{MPESMB, TAPCGPG}, other geometry adaptation methods in SMB, e.g., \cite{POLYMORPH, TCBPGS, EPCSMBGE, MAIC}, are also subject to the same limitations. By making use of CPs, we propose a method that tends to overcome such limitations in content adaptation.

\section{Learning Constructive Primitives}
In this section, we first describe the motivation underlying our approach and then present our approach to producing constructive primitives (CPs) in SMB.

\subsection{Motivation}

\par
As summarized in Sect. I, there are three non-trivial issues in existing SMB level generators: game quality assurance, generation efficiency, and procedural level controllability.

\par
By a close look at existing SMB level generators reviewed in Sect. II, we observe that the content space on all the complete procedural levels is huge. As there are an enormous variety of combinations among game elements and structures at procedural levels, an approach working on such content space inevitably faces a greater challenge in managing quality assurance and efficiency in PCG. Thus, controlling the level generation may be rather complicated and difficult \cite{AGACMPCVGL}. Nevertheless, a complete procedural level in SMB can be decomposed into a number of segments as evident in segment-based level generators \cite{LAUNCHPAD}. Thus, partitioning a procedural level into fixed-size game segments without relying on any concepts, e.g., rhythm, allows us to explore the SMB content space from an alternative perspective. As a result, all the possible segments form a new content space of lower complexity. We believe that it is less difficult to understand the properties of short game segments and hence the use of those segments as building blocks would facilitate tackling three non-trivial issues in SMB.

\par
For quality assurance, there are generally two methodologies in developing such a mechanism in PCG \cite{SBPCG,LBPCG}: deductive vs. inductive. To adopt the deductive methodology, game developers have to understand the content space fully and know how to formulate/encode their knowledge into rules, fitness or constraints explicitly. In the presence of a huge content space, however, it would be extremely difficult to understand the entire content space so that either a small region of a content space is merely taken into account or less accurate (even conflicted) rules/constraints have to be used in PCG. The former could significantly limit the number of games generated by PCG while the latter is responsible for low quality games generated by PCG. Nevertheless, we observe that some rules/constraints are easy to design/identify while a complete set of rules for evaluating the content quality are hard to handcraft. For example, overlapped tubes in SMB is unacceptable and can be easily detected with a simple rule. On the other hand, a learning-based PCG (LBPCG) framework \cite{LBPCG} was recently proposed where an inductive methodology, i.e. learning from data, was advocated for quality assurance. As game content is observable but less explainable, it is easier for game developers to make a judgement on quality for a specific game by applying their knowledge implicitly than to encode their knowledge into rules or constraints. Thus, the LBPCG suggests that a quality evaluation function should be learned from data annotated by game developers. Nevertheless, the annotation for producing learning examples may be more time-consuming than designing simple rules to detect those apparent low quality games. Hence, a hybrid approach to quality assurance would allow us to exploit the synergy between rule-based and learning-based methods.

\par
\begin{figure}[htp]
\small
\centering
\includegraphics[width=9cm]{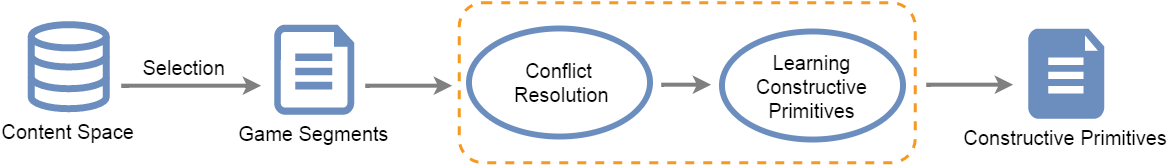}
\caption{The constructive primitive learning process for SMB.}
\label{fig:method}
\end{figure}

With the motivation described above, we propose a hybrid approach to producing CPs, quality yet controllable game segments, in SMB. Fig.~\ref{fig:method} illustrates  the main steps of our approach. First of all, game developers choose a region of interest from the entire content space via control parameters. Then game segments in the region of interest are evaluated by a set of easy-to-design handcrafted conflict resolution rules and the subsequent data-driven quality evaluation function that deals with more complicated quality issues. Survivals of game segments become CPs.

\subsection{Content Space}

\begin{figure}[hp]
\small
\centering
\includegraphics[width=9cm]{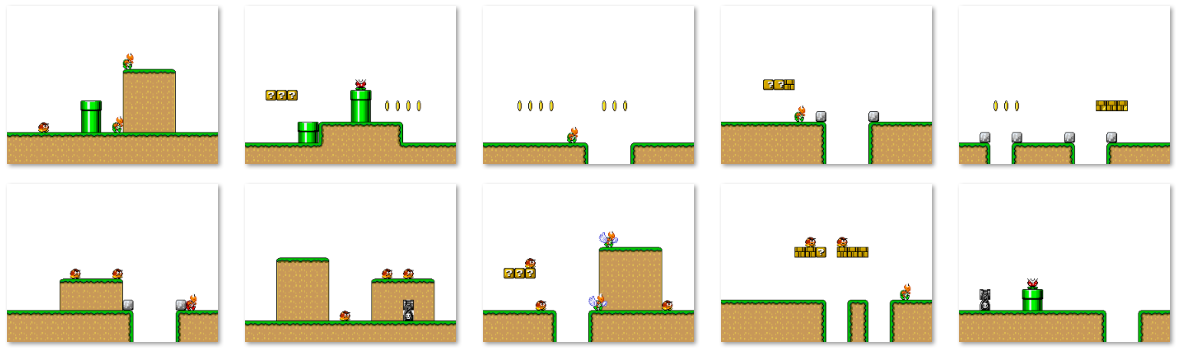}
\caption{Game segment instances in our content space.}
\label{fig:gc}
\end{figure}

Based on our observation from empirical studies, it is sufficient to cover rich yet diverse types of games by using a game segment of 20 in length and 15 in height. Some typical game segment instances are illustrated in Fig.~\ref{fig:gc}.

\par
The SMB content is naturally specified by a 2D grid similar to an image. However, this leads to a 300-dimensional content space in our case where there are a lot of redundancy, e.g., the uniform background. Motivated by the previous work \cite{AGACMPCVGL}, we employ a list of design elements as our content space representation  where a design element refers to an atomic unit used in a procedural level generation \cite{SBPCG,AGACMPCVGL}, e.g., enemy, boxes, coins, cannon, gap and so on. By using this representation, we can not only specify the content space concisely but also gain the direct controllability on low-level content features, e.g., coordinates of enemies and coins. As listed in Table~\ref{table:FCS}, 85 controllable features are employed in our representation. Such representation is similar to the previous work \cite{AGACMPCVGL}. In our content space, the design elements in each type are sorted in decreasing order along $x$ dimension.

\begin{table*}[htbp]
\caption{Design elements used in representing the content space.}
\centering
\begin{tabular}{ l | l | l | l  }
\hline
\hline
\textbf{ID} & \textbf{Description} & \textbf{ID} & \textbf{Description}   \\ [0.5ex]
\hline
1 & $height$ of initial platform  &42 - 47 &  $x$, $y$, $height$, $w_{before}$, $w_{after}$ and $type$ of the second tube  \\  \hline
2 & number of gaps &48 - 53 & $x$, $y$, $height$, $w_{before}$, $w_{after}$ and $type$ of the third tube   \\  \hline
3 - 5 & $x$, $width$ and $type$ of the first gap &54 & number of boxes    \\  \hline
6 - 8 & $x$, $width$ and $type$ of the second gap &55 - 58 &  $x$, $y$, $width$ and $type$ of the first boxes  \\  \hline
9 - 11& $x$, $width$ and $type$ of the third gap &59 - 62 &  $x$, $y$, $width$ and $type$ of the second boxes   \\  \hline
12 & number of hills &63 &  number of enemies   \\  \hline
13 - 15 & $x$, $width$ and $height$ of the first hill &64 - 66 &  $x$, $y$ and $type$ of the first enemy   \\  \hline
16 - 18 & $x$, $width$ and $height$ of the second hill &67 - 69 &  $x$, $y$ and $type$ of the second enemy   \\  \hline
19 & number of cannons &70 - 72 &  $x$, $y$ and $type$ of the third enemy   \\  \hline
20 - 24 & $x$, $y$, $height$, $w_{before}$ and $w_{after}$ of the first cannon & 73 - 75 &   $x$, $y$ and $type$ of the forth enemy  \\  \hline
25 - 29 & $x$, $y$, $height$, $w_{before}$ and $w_{after}$ of the second cannon &76 - 78 & $x$, $y$ and $type$ of the fifth enemy    \\  \hline
30 - 34 & $x$, $y$, $height$, $w_{before}$ and $w_{after}$ of the third cannon &79 &  number of coins   \\  \hline
35 & number of tubes &80 - 82 &  $x$, $y$ and $width$ of the first coins   \\  \hline
36 - 41 & $x$, $y$, $height$, $w_{before}$, $w_{after}$ and $type$ of the first tube &83 - 85 & $x$, $y$ and $width$ of the second coins  \\  \hline
\end{tabular}
\label{table:FCS}
\end{table*}

\par
While design element parameters in Table~\ref{table:FCS} have a wide range that specifies the entire content space, we confine our concerned content space to a non-trivial region of the entire content space by setting the maximum number of gaps, hills, tubes, cannons, boxes, coins and enemies appeared in a game segment are 3, 2, 3, 3, 2, 2 and 5 respectively. Consequently, there are roughly $9.72\times10^{37}$ game segments in our content space. This content space should be sufficient for generating content with a variety of geometrical features, level structures and difficulties required by SMB.

\subsection{Conflict Resolution}

\par
 In our content space, there are quite a number of game segments that contain conflicting design elements. For instance, ``\dots \emph{Tube(7,2,3,4,4,flower)}\dots\emph{Cannon(7,3,1,3,2)}\dots" represents a game segment of at least one tube and one cannon but their $x$ coordinates are same. Thus, the cannon and tube are overlapped together and this conflicting situation makes the segment aesthetically unappealing.
\par
Motivated by the previous work \cite{ELSMBGE, AGACMPCVGL}, we adapt a class of rules presented in \cite{AGACMPCVGL} for our requirement.  As a result, the conflict resolution rules effectively discard those game segments of geometrically overlapped design elements including gap, enemy, tube, cannon, boxes and coins.

\subsection{Learning Constructive Primitives}

\par
After filtering out those obviously unappealing game segments, the tailored content space still contains a lot of low quality segments, e.g., segments of unreachable coins and boxes, segments of being too difficult/easy to play, segments of unbalanced resources and aesthetically unappealing structures. Inspired by the LBPCG work \cite{LBPCG}, we would learn a quality evaluation function from annotated game segments to remove unplayable/unacceptable segments. To carry out this idea, a binary classifier is trained where its input is the 85D feature vector of a game segment and its output is a binary label that predicts the quality of a game segment. Game segments labeled as positive are CPs and would be used for online level generation and content adaptation described in Sect. IV.
\par
\begin{figure}[htp]
\small
\centering
\includegraphics[width=9cm]{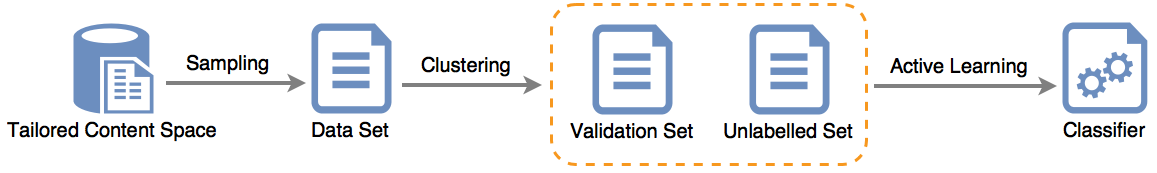}
\caption{The constructive primitive learning process.}
\label{fig:classifier}
\end{figure}

To establish a data-driven evaluation function, training examples are required but have to be provided from game developers. As the tailored content space is still huge, it is infeasible to annotate all possible games in this content space. To keep the content space manageable, a proper sampling can be applied to achieve a much smaller data set of the same properties as the content space. Motivated by the success in the LBPCG work \cite{LBPCG}, we conduct clustering analysis on the data set and further employ active learning based on the clustering results to minimize a game developer's efforts in data annotation. In summary, this CP learning process is depicted in Fig.~\ref{fig:classifier}.

\subsubsection{Sampling}
\par
For sampling, we apply the simple random sampling with replacement \cite{SP}, an unbiased sampling technique, to the tailored content space for a manageable data set. As a result, we randomly set all the controllable features in the tailored content space to ensure that each game segment in this content space has the equal probability to be selected. The size of data set is determined via the sample size determination (SSD) algorithm suggested in \cite{SP}.
With the theoretical justification, the SSD can decide the size of a sampled data set without loss of non-trivial information. By applying the SSD to our tailored content space, it is suggested that a data set of 19,000 games should be sufficient.

\subsubsection{Clustering}
\par
We apply the CURE algorithm \cite{CURE} on the sampled game set for clustering analysis since this hierarchical clustering algorithm can deal with a large data set and discover the clusters of different sizes in complex shapes. There are four parameters in CURE algorithm: the number of clusters, sampling rate, shrink factor and the number of representative points. By using the dendrogram tree achieved, the number of clusters is automatically decided based on the longest k-cluster lifetime \cite{CMCUEA}. The rest of parameters are set to defaults suggested in \cite{CURE}; i.e., 2.5\% for sampling rate, 0.5 for shrink factor and 10 representative points, respectively. Due to the existence of two different feature types, i.e. nominal and ordinal, we employ the mixed-variable distance metric \cite{DM} in the CURE. After clustering, we found 106 clusters from this sampled data set, and the clustering results would be used to facilitate active learning.

\subsubsection{Active Learning}
\par
For binary classification, there are two error types: \emph{false negative} (type-I error) where a high quality segment is misclassified as low quality and \emph{false positive} (type-II error) where a low quality segment is  misclassified as high quality. Obviously, a type-II error could result in a catastrophic effect while a type-I error simply shrinks the content space slightly. As a result, we formulate our classification as a cost-sensitive learning problem where the type-II error incurs a higher cost. By looking into several state-of-the-art classification techniques, we found that the weighted random forests(WRFs) \cite{URFLID}, a cost-sensitive oblique random forests \cite{RF} classifier, fully meet our requirements for active learning. In our work, the parameters of WRFs \cite{URFLID} are set via validation as follows: 2:1, 50, 5, 10 and 9 for the cost ratio, the number of trees, the number of combined features, the number of feature groups selected at each node, and depth of trees, respectively.
\par
After clustering, a small number of segments are selected from each cluster to form a validation set. The number of segments selected from each cluster is proportional to the cluster size. Totally, there are 800 segments in the validation set. We annotate each game in the validation set by visual inspection in order to evaluate the generalization performance of a classifier during active learning.
\par
During active learning, we randomly choose 100 segments and annotate them visual inspection to train the initial WRFs. In each iteration, we find 100 segments of the highest uncertainty scores, defined by
$s_i = 1 - P(\hat y|x_i)$ where $\hat y$ is the predicted label of segment $x_i$, and $P(\hat y|x_i)$ is the probability of this prediction, and annotated them to be examples for re-training WRFs. The active learning stops when the accuracy of WRFs on the validation set no longer increases. Our active learning algorithm is summarized in Algorithm~\ref{algo:AWORF}.

\begin{algorithm}[t]
\caption{Active Constructive Primitive Learning}
\label{algo:AWORF}
\begin{algorithmic}
\STATE \textbf{Input:} Sampled data set $U$ and clustering results on $U$.
\STATE \textbf{Output:} WRFs binary classifier.
\STATE \textbf{Initialization:}
 Based on the clustering analysis results, create a validation set $V$ of 800 examples.
 \STATE \textbf{Active Learning:}
\STATE Annotate 100 segments randomly selected from $U$ via visual inspection to form a training set $L$. Train WRFs on $L$ to obtain an initial binary classifier.
\REPEAT
\FORALL{$x_i\in U$}
\STATE Label $x_i$ with the current WRFs.
\STATE Calculate the uncertainty score $s_i$ of $x_i$.
\ENDFOR
\STATE Annotate 100 segments of the highest uncertainty score in $U$
to form a new training set $L$.
\STATE Re-train the WRFs with the examples in $L$.
\UNTIL{The overall accuracy on $V$ does not increase.}
\end{algorithmic}
\end{algorithm}

\par
Once the learning-based evaluation function is constructed, we use it along with the aforementioned rules to produce CPs of favorable properties that ensures to gain the direct control of CPs via the relevant design elements in terms of geometrical features, level structures and difficulties.

\section{Online Generation and Real-time Adaptation}
In this section, we come up with the techniques in applying constructive primitives to online procedural level generation and real-time content adaptation in terms of DDA.

\subsection{Online Game Generation}

\par
As described in Sect. III, CPs provide quality building blocks and hence lumping them together can easily lead to a procedural level of aesthetically appealing content with a path between entrance and exit. In SMB, there are a variety of procedural levels that can be categorized based on a number of properties, e.g., density \cite{ELSMBGE}, leniency \cite{AERLG} and linearity \cite{AERLG}. As our CPs are represented by design elements, we can generate a procedural level of pre-setting property via controllable level generation parameters.
\par
Motivated by the previous works \cite{ELSMBGE,AERLG}, we employ three controllable level generation parameters, i.e., density, leniency  and linearity, to generate a variety of levels online. The density controls the complexity of geometrical structures, e.g., a high density leads to many overlapping hills. The leniency decides the level difficulty in gameplay; intuitively, a high leniency results in an easy-to-play level.  The linearity is yet another parameter that ensures there is a linear structure in a generated level; a large value leads to a level of highly linear structures. Each level generation parameter is carried out by setting the proper values to relevant design elements in CPs as follows:
\begin{itemize}
\item \emph{Leniency}. This parameter is implemented via controlling the number and type of enemies, number and width of gaps, number of cannons and tube flowers in CPs.
\item \emph{Density}. This parameter is decided by the number and coordinates of hills in CPs.
\item \emph{Linearity}. This parameter operates by specifying the height of platform, the number of hills, $y$ coordinates of tubes and cannons in CPs.
\end{itemize}

Each level generation parameter is set to $\{1,2,3\}$, which divides our CPs into three categories reflecting the different properties specified by a specific parameter.

\begin{algorithm}[htp]
\caption{Online Procedural Level Generation}
\label{algo:GGA}
\begin{algorithmic}
\STATE \textbf{Input:} Level generation parameter setting and the number of CPs (specifying the level length to be generated).
\STATE \textbf{Output:} Procedural level of a fixed length specified by the number of CPs.
\STATE \textbf{Generation:}
\STATE Generate an empty segment as the game entry.
\REPEAT
\STATE Randomly produce a CP of the specified $leniency$, $density$ and $linearity$ parameter values.
\STATE Append this CP to the existing procedural level.
\UNTIL{The level contains the specified number of CPs.}
\STATE Generate the exit for this procedural level.
\end{algorithmic}
\end{algorithm}

\par
To generate a complete level, we first specify the desired values to level generation parameters that fix the parameter values of relevant design elements and set other irrelevant design elements in CPs randomly. Thus, an iterative process is undertaken by merging the CPs of the specified properties together until reaching a pre-specified length or the number of CPs pre-specified. As each CP is produced very efficiently, our level generator works online. The online level generation algorithm is summarized in Algorithm~\ref{algo:GGA}.
It is worth stating that linearity may conflict with density. Hence, we stipulate that the value set to density overrides that of linearity for their shared design elements. In other words, we do not want to  generate highly linear procedural levels, which is often considered
as aesthetically unappealing.

\subsection{Real-time Content Adaptation}

\par
For content adaptation, we confine our work on CPs to only the dynamic difficulty adjustment (DDA) where the game difficulty is automatically adjusted according to a player's performance.
\par
For DDA, it is essential to define content difficulty and measure a player's performance. In our work, a difficulty parameter of five levels is defined for each CP based on its relevant design elements that affect a player's performance, such as the number and the type of enemies, the number and the width of gaps, the number of cannons and tube flowers. For instance, a CP of the highest difficulty level contains as least two enemies, one cannon and one gap while there are at most one Goomba and one flower tube in a CP of the lowest difficulty level. Measuring a player's performance may be complicate if all the aspects need considering. In our work, we simplify this by only taking the survival rate on CPs into account.
\par
By means of the segment-based methodology, we would make use of our CPs for real-time DDA, i.e., the performance is measured locally on a CP instead of a level and the DDA is done instantly in response to a player's local performance. This is naturally a sequential decision process and we would formulate it as a Markovian decision process (MDP) \cite{MDP}. Thus, our goal is to find an optimal policy that maximizes the expected long-term performance via adjusting the difficulty of generated CPs. To attain this goal, we define a regret $\rho$ at time $T$ (i.e., after $T$ CPs have been played) as the absolute difference between an expected survival rate and the expectation of rewards:
\begin{equation}
\label{regret}
\rho = |\theta_{opt} - \frac{1}{T}\mathbb{E}[\sum_{t=1}^{T}r_{t}]|,
\end{equation}
where $\theta_{opt}$ is a pre-set optimal survival rate and $r_t$ is a binary reward: $r_t=1$ if the player survives and $0$ otherwise. For instance, we set $\theta_{opt} = 0.8$ if the performance of the survival rate 0.8 is expected. For a given $\theta_{opt}$, levels of proper difficulties can be generated for a player to gain the expected performance. Thus, minimizing this regret is key to our content adaptation.
To solve this optimization problem, we employ the Thompson sampling \cite{TS}, an effective and efficient heuristic method used in binary reward case in MDP.
\par
Let $\theta_i$ denote a player's survival rate of difficulty level $i$ ($i=1,\cdots,5$). Thus, a player survives with the probability $\theta_i$ and reciprocally dies with the probability $1-\theta_i$ when they play a CP of difficulty level $i$. During gameplay, one plays a number of CPs of different difficulty levels sequentially. When a CP of difficulty level $a_t=i$ is completed at time $t$, a binary reward is assigned. Given the player's survival rate $\theta_{i}$ corresponding to $a_t$, the reward likelihood is
\begin{equation}
\label{likelihood}
P(r_{t}|a_t = i, \theta_{i}) = \theta_{i}^{r_{t}}(1-\theta_{i})^{1-r_{t}}.
\end{equation}
Furthermore, let $D_T=  ( a_t,r_t )_{i=1}^T$ denote the historical profile regarding corresponding difficulties and rewards after $T$ CPs have been played.
By using a conjugate prior, $\theta_i\sim Beta(1,1)$, the posterior distribution of survival rate based on the likelihood in Eq. (\ref{likelihood}) is $P(\theta_i|D_T) \propto \prod_{t=1}^T P(r_{t}|a_t,\theta_i)P(\theta_i)$, which leads to $\theta_i|D_T \sim Beta(\alpha_i\!+\!1,\beta_i\!+\!1)$  where $\alpha_i$ and $\beta_i$ are the number of survives and deaths when playing the CPs of difficulty level $i$ in $D_T$.

\par
For content adaptation, we follow the typical setting in real SMB games: at the beginning, a player is put in small state, i.e., the weakest form of Mario, where the player is not allowed to use powerful weapon (e.g., throwing fireballs) and then turns into other states by powering up with a mushroom or fire flower. Based on the gameplay information recorded in $D_T$, CPs are randomly produced according to
$Beta(\alpha_i\!+\!1,\beta_i\!+\!1)$ and the CP of difficulty level $i$ is chosen as a game segment to play if it results in the least regret defined in Eq. (\ref{regret}). After a CP of difficulty level $i$ is played, the posterior probability $Beta(\alpha_i\!+\!1,\beta_i\!+\!1)$ is updated based on the performance on the CP. Thus, this content adaptation process continues until a player quits, as summarized in Algorithm~\ref{algo:TSGA}. It is worth stating that this algorithm is presented for a life-long gameplay scenario but easily adapted for multiple gameplay episodes by substituting the Beta conjugate prior, $Beta(1,1)$, with the posterior obtained from the last episode,
$Beta(\alpha_i,\beta_i)$, in the initialization whenever a new episode starts.

\begin{algorithm}[htp]
\caption{Real-time Content Adaptation for DDA}
\label{algo:TSGA}
\begin{algorithmic}
\STATE \textbf{Input:} Optimal survival rate $\theta_{opt}$.
\STATE \textbf{Initialization:} $t\! \leftarrow \! 0$,  $\alpha_i \! \leftarrow \! 1$ and $\beta_i\! \leftarrow \! 1$ for $i=1,\cdots,5$.
\REPEAT
\STATE $t \! \leftarrow \! t+1$.
\IF{$t \!==\! 1$ or $r_{t-1}==0$}
\STATE Sample CPs according to $\theta_i \sim Beta(\alpha_i,\beta_i)$ \\
for $i=1,\cdots,5$.
\STATE Choose the CP of $a_t = \argmin_i |\theta_{opt} - \theta_i|$.
\ELSE
\STATE Sample CPs according to $\theta_i \sim Beta(\alpha_i,\beta_i)$ \\
for  $i \!=\! \max(1,a_{t-1}\!-\!1)$, $a_{t-1}$ and $\min(a_{t-1}\!+\!1,5)$.
\STATE Choose the CP of $a_t = \argmin_i |\theta_{opt} - \theta_i|$.
\ENDIF
\STATE Generate the chosen CP of $a_t$.
\IF{$r_t==1$ }
\STATE $\alpha_{a_t} \! \leftarrow \! \alpha_{a_t} + 1$.  //the player survives
\ELSE
\STATE $\beta_{a_t} \! \leftarrow \! \beta_{a_t} + 1$.  //the player dies and a new game starts
\ENDIF
\UNTIL{Gameplay stops.}
\end{algorithmic}
\end{algorithm}

\section{Experimental Results}
In this section we report results in the CP learning, online level generation and simulated DDA. The game engine adopted in our experiments is a modified version of the open-source \emph{Infinite Mario Bros} used in the \emph{Mario AI Championship} \cite{MAIC, TMAC}. Our level generators that yield results reported in this section are publicly available on our project website\footnote{\url{http://staff.cs.manchester.ac.uk/\~shipa/mario.html}}.

\subsection{Results on Constructive Primitive Learning}

\begin{table*}[htbp]
\caption{Top 30 important design elements in the constructive primitive learning.}
\centering
\begin{tabular}{ l | l | l | l | l | l }
\hline
\hline
Rank & Description & Rank & Description & Rank & Description \\ [0.5ex]
\hline
1 & $x$ coordinate of the first gap &11 & number of enemies & 21 & $y$ coordinate of the first cannon \\  \hline
2 & $width$ of the first hill &  12 & number of cannons &  22 &  $y$ coordinate of the first box\\  \hline
3 & $x$ coordinate of the first enemy & 13 & $x$ coordinate of the first cannon&   23 & $width$ of the first box \\  \hline
4 & $height$ of the first hill &  14 & $y$ coordinate of the first enemy&  24 & $type$ of the third enemy \\  \hline
5 & $width$ of the first gap & 15 & $width$ of the second hill &25 & $x$ coordinate of the first coin \\  \hline
6 & $height$ of the first cannon & 16 & $w_{before}$ of the first tube&26 & $width$ of the second gap  \\  \hline
7 & number of gaps & 17 & $y$ coordinate of the first coin & 27 & $height$ of the second cannon \\  \hline
8 & $w_{before}$ of the first cannon & 18 & $height$ of initial platform &28 & $w_{before}$ of the second cannon  \\  \hline
9 & type of the first enemy & 19 & $x$ coordinate of the second gap & 29 & $y$ coordinate of the third cannon \\  \hline
10 & $w_{after}$ of the first cannon & 20 &  number of hills &30 &  $x$ coordinate of the first tube  \\  \hline
\end{tabular}
\label{table:RSF}
\end{table*}

\begin{figure}[htbp]
\small
\centering
\includegraphics[width=8cm]{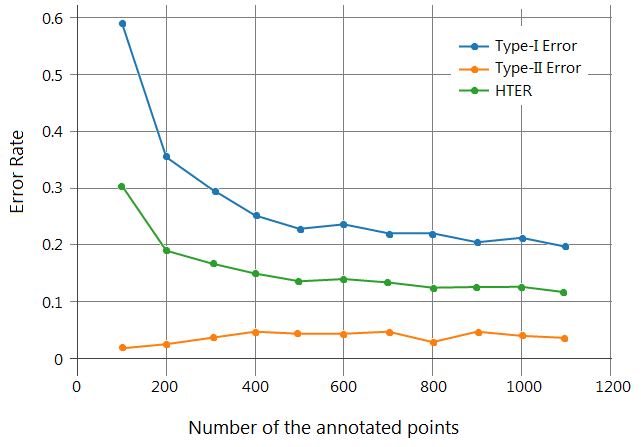}
\caption{Performance evolution on the validation set during active learning.}
\label{fig:err}
\end{figure}

Based on the learning algorithm described in Sect. III, Fig.~\ref{fig:err} illustrates the evolutionary performance of our active learning on the validation set, including types I and II error rates as well as their average, the \emph{half total error rate} (HTER). From Fig.~\ref{fig:err}, it is observed that the active learning converges after 1100 data points.

\begin{figure}[htbp]
\small
\centering
\includegraphics[width=9cm]{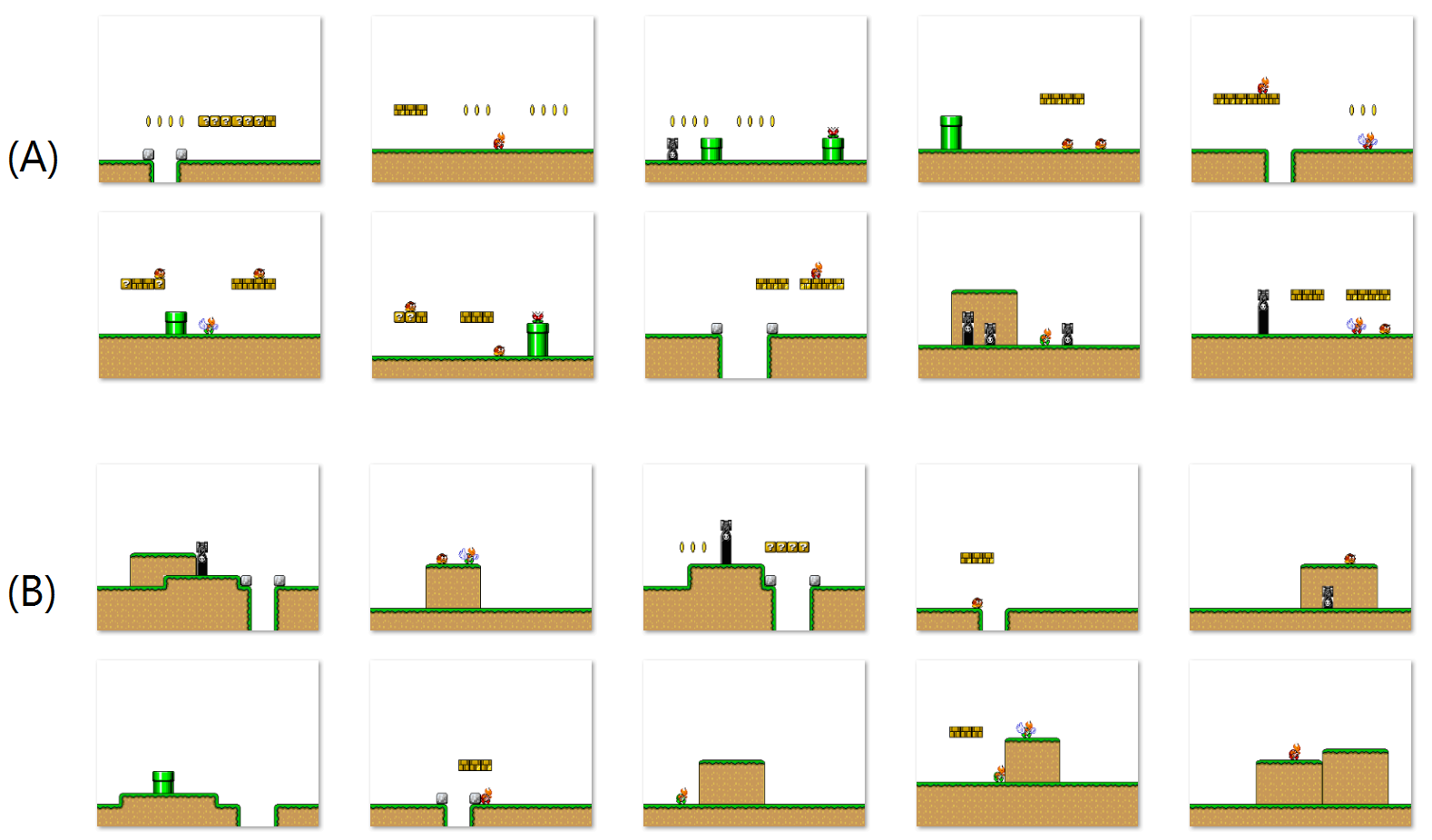}
\caption{Misclassified segment instances. (A) Segments leading to  type I error. (B) Segments leading to type II error.}
\label{fig:type12}
\end{figure}

While the final HTER is around 11.67\%, the corresponding type-I error rate is around 19.66\% and some misclassified segment instance are shown in Fig.~\ref{fig:type12}(A).
By analyzing the clustering results, we find out that
those segments are concentrated in a quite small region in the content space. Therefore, they are unlikely to be sampled so as to missed in establishing the validation set.
It is evident from  Fig.~\ref{fig:err} that our cost-sensitive classifier performs well in minimizing the type-II error; the type-II error rate is approximately 3.69\%. Among those low quality segments misclassified as high quality, about 0.74\% segments contain unreachable resources (e.g., coins and boxes) and the rest segments are either too easy/difficult to play or less appealing aesthetically as exemplified in Fig.~\ref{fig:type12}(B).
While such segments may result in negative gameplay experience, fortunately, none of unplayable segments in the validation set was misclassified.

In this experiment, we use the permutation test pertaining to RFs \cite{RF} to measure the importance of design elements in the CP learning. As a result, Table~\ref{table:RSF} list the top 30 design elements that significant affect the quality of CPs. From Table~\ref{table:RSF}, it is observed that there is a close connection between the design element such as gap/hill/cannon and quality of a game segment. In addition, $x$ and $y$ coordinates and geometrical properties of game elements (e.g., their width and height) are among the most important features for the CP prediction.

\subsection{Results on Online Level Generation}

\par
In all our experiments, a game level are confined to a 2D map of 200 in length and 15 in height, as same as the setting in previous works, e.g., \cite{LAUNCHPAD, AGACMPCVGL, ELSMBGE, AMLLG, PLGORE}. We evaluate our online level generator in terms of expressive range, controllability and generation efficiency.

\subsubsection{Expressive Range}
\par
Expressive range refers to the range and variation of procedural levels according to an evaluation metric \cite{AERLG} that measures the property of procedural levels. By using a specific metric, the expressive range is often used to visualize the space of all possible procedural games.  For evaluation, we use \emph{linearity} \cite{AERLG}, \emph{density} \cite{ELSMBGE}, \emph{leniency} \cite{AERLG} and \emph{compression distance} \cite{ELSMBGE} as our metrics. Such metrics allows us to reveal global properties of game levels generated by a level generator.
\par
For linearity, we use the method suggested in \cite{ACEPLGM} to find a line that fits the profile of a procedural level, and the coefficient of determination $r^2$ is used to estimate the degree of linearity. For density, we count the number of all possible standing positions in a game level \cite{ELSMBGE}. For leniency, we assign a value to each type of game elements as same as used in \cite{LAUNCHPAD, ELSMBGE} (i.e., enemy: -1.0, gap, cannon or flower tube: -0.5, and powerup: +1.0). The overall leniency score is the sum of the three values. Finally, the normalized compression distance (NCD) \cite{ELSMBGE} is employed to measure the dissimilarity between two game levels; the higher this distance the more dissimilar two levels.
\par
For a thorough evaluation, we compare our level generator with a number of typical SMB level generators reviewed in Sect. II, including Notch \cite{NOTCH}, Parameterized Notch (PN) \cite{FAMGCQ}, Grammar Evolution (GE) \cite{ELSMBGE} and  Launchpad generators \cite{LAUNCHPAD}. For fairness, we adopt the experimental protocols suggested in \cite{ACEPLGM}. As a result, each of level generators generates 200 procedural levels for evaluation in terms of linearity, density and leniency\footnote{The levels generated by those generators were provided by N. Shaker and only 200 procedural level images are available.} and the NCD metric is applied to 200 pairs of levels. The level generation parameters used in ours are set randomly and others use their default settings. To our knowledge, the parameters in PN and Launchpad were set randomly, and there is no parameter in GE.
The scores measured with a metric are normalized to the range of [0,1].

\begin{figure}[htbp]
\small
\centering
\includegraphics[width=9cm]{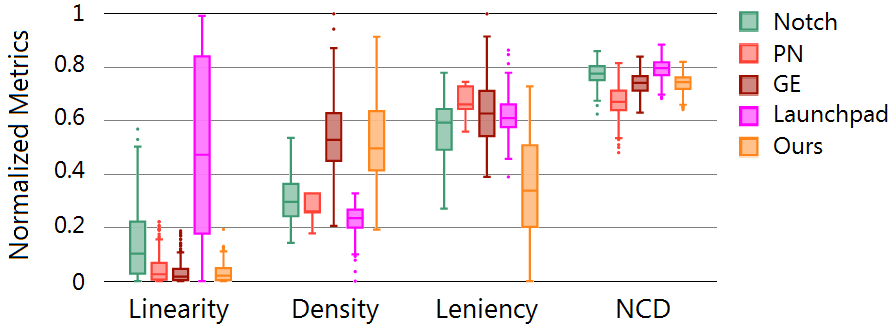}
\caption{Express ranges of different SMB level generators.}
\label{fig:comp}
\end{figure}

\par
The results on expressive ranges are shown in Fig.~\ref{fig:comp}. It is observed from Fig.~\ref{fig:comp} that Launchpad can generate both linear and nonlinear levels while the rest generators tend to generate non-linear levels. As described in Sect. IV.A, the setting in our generator prefers nonlinear levels. Regarding density, the expressive ranges of GE and ours are wider than others. However, others tend to generate levels with a low density. Regarding leniency, Launchpad and PN tend to generate levels of medium difficulty levels. In contrast, others including ours may generate both difficult and easy game levels. Among all the generators, the expressive range of ours is widest in terms of leniency and density. The expressive range differences among those level generators account for the use of different content spaces and level generation parameters. From Fig.~\ref{fig:comp}, it is evident that PN receives the lowest NCD score, which implies relatively similar levels are generated. In contrast, others generate levels of a greater diversity as there are larger NCD distances among levels they generate, and ours generates levels of medium diversity in comparison to others.

\subsubsection{Controllability}
\par
\begin{figure}[htbp]
\small
\centering
\includegraphics[width=8.5cm]{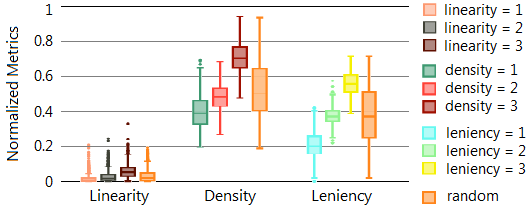}
\caption{Expressive ranges of our online level generator corresponding to different controllable parameter values.}
\label{fig:cv}
\end{figure}

We further evaluate the controllability of our level generator. In general, controllability can be reflected in the expressive ranges of procedural levels generated with different level generation parameter settings \cite{LAUNCHPAD}. As ours have three level generation parameters and each may take one of three values as described in Sect. IV.A, we exhaustedly generate nine sets of levels by fixing one parameter with a specific value and randomly setting all other parameters each time. To see the controlling effect clearly, we also generate a set of levels by setting all the parameters randomly. Thus, we achieve 10 level sets where each contains 1000 levels for reliability. In terms of linearity, density and leniency, the expressive ranges of levels controlled by different parameters are shown in Fig.~\ref{fig:cv} where it is clearly seen that the levels of a specific property are generated by properly controlling a parameter.

\par
\begin{figure*}[htbp!]
\small
\centering
\includegraphics[width=18cm]{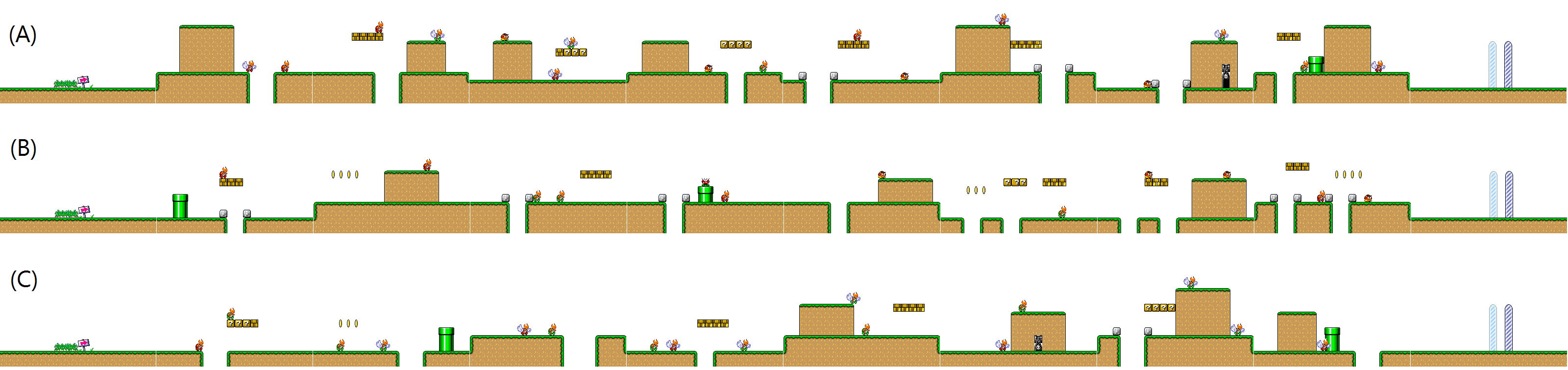}
\caption{Exemplar levels levels with different linearity values.  (A) \emph{linearity} = 1. (B) \emph{linearity} = 2. (C) \emph{linearity} = 3.}
\label{fig:l3}
\end{figure*}
\begin{figure*}[htbp!]
\small
\centering
\includegraphics[width=18cm]{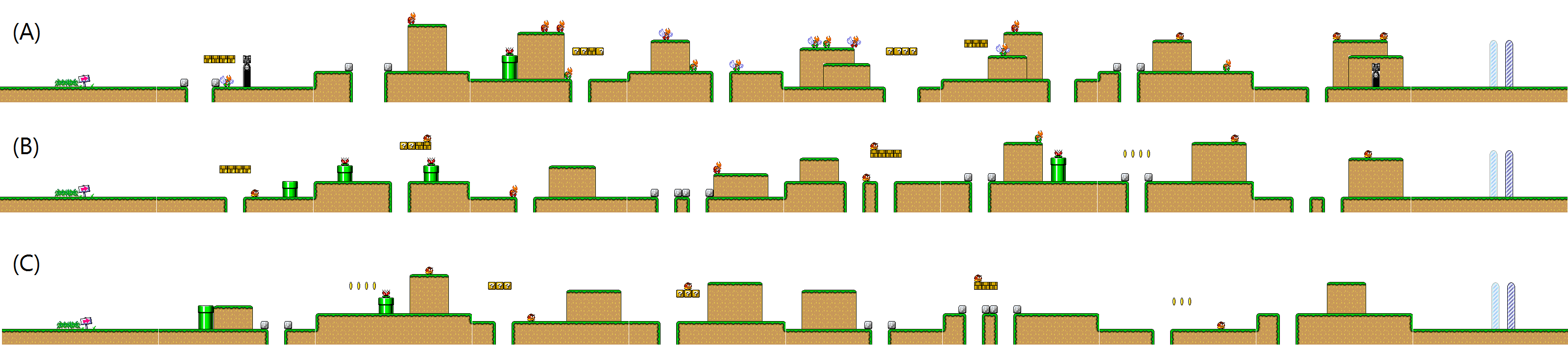}
\caption{Exemplar levels levels with different leniency values. (A) \emph{leniency} = 1. (B) \emph{leniency} = 2. (C) \emph{leniency} = 3.}
\label{fig:e3}
\end{figure*}
\begin{figure*}[htbp!]
\small
\centering
\includegraphics[width=18cm]{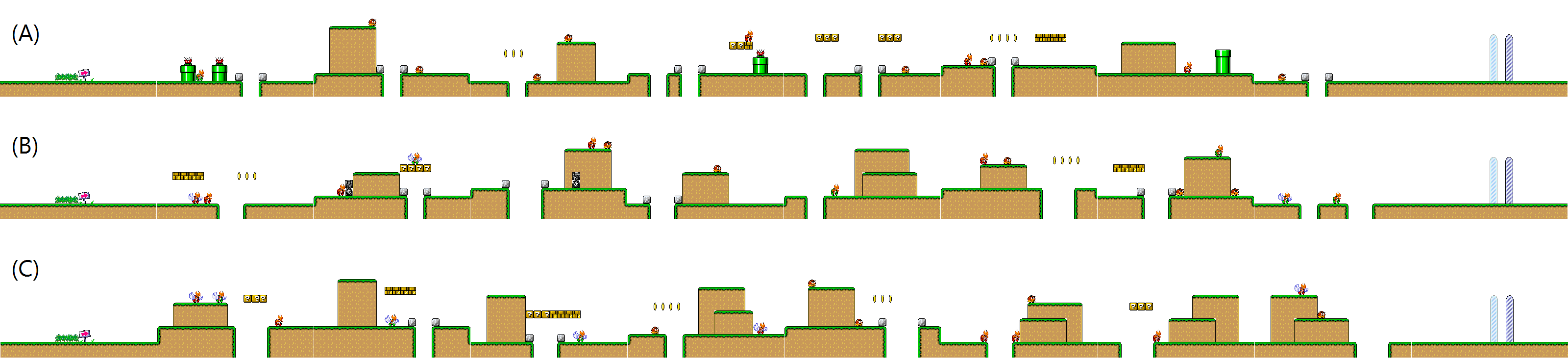}
\caption{Exemplar levels generated with different density values. (A) \emph{density} = 1. (B) \emph{density} = 2. (C) \emph{density} = 3.}
\label{fig:d3}
\end{figure*}

For exemplification, Fig.~\ref{fig:l3} - \ref{fig:d3} illustrate some levels generated by controlling parameters in a specific way. By visual inspection, we observe that the level shown in Fig.~\ref{fig:l3} (C) is smoother than those in Fig.~\ref{fig:l3} (A) and (B) since it is generated by using the highest linearity value. From Fig.~\ref{fig:e3}, it is seen that the profile shown in Fig.~\ref{fig:e3} (C) this level is easier to play (e.g., fewer enemies and narrower gaps) than those shown in Fig.~\ref{fig:e3} (C) due to the use of the highest leniency value. It is evident from Fig.~\ref{fig:d3} that there are more complicated geometrical structures (e.g., more overlapping hills) in the level shown in Fig.~\ref{fig:d3} (C) than those shown in Fig.~\ref{fig:d3} (A) and (B) as the highest density value is applied.

\subsubsection{Efficiency}
\par
Generation efficiency is often evaluated by the actual time taken in a level generation. As we do not have codes of level generators used for comparison in our experiments, we can only report the result on ours. By testing on a PC (Intel Core i5-3470 processor with 8GB memory), our level generator takes only 0.057 sec on average to generate a procedural level, $200 \! \times \! 15$ 2D map, which should be able to meet the online generation requirements.

\subsection{Results on Real-time Content Adaptation}

\par
For the evaluation of our proposed real-time content adaptation for DDA, we conduct simulations with sophisticated Mario controllers of different types instead of human players as suggested in \cite{TAPCGPG}. The use of agents for DDA test may benefit from stable agents' behavior, their diversified playing styles and a wide range of skills \cite{TAPCGPG}. As a result, we employ 15 agents submitted to the Gameplay track of the Mario AI Competition \cite{TMAC}. To test our adaptive generator, we use \emph{completion rate}, the ratio of the actual distance travelled over to the length of a game level being played, as a evaluation criterion for DDA \cite{TMAC}. Moreover, we further examine the online learning performance of our adaptation algorithm based on the completion rates of three typical agents in response to adaptive content generated for an optimal survival rate, $\theta_{opt}$. In our simulation, a level generated with our adaptation algorithm is limited to a maximum length of 200.
\par
For reliability, each of 15 agent played on three sets of adaptive games generated by Algorithm 3 with different optimal survival rates,  $\theta_{opt}=0.80,0.87,0.95$, where each set consists of 200 levels.
For a baseline, we also randomly generated 200 levels of the same length and refer them to \emph{static} games as no DDA is applied and each agent also plays the games in the static game set.
\par
\begin{figure*}[htbp]
\small
\centering
\includegraphics[width=18cm]{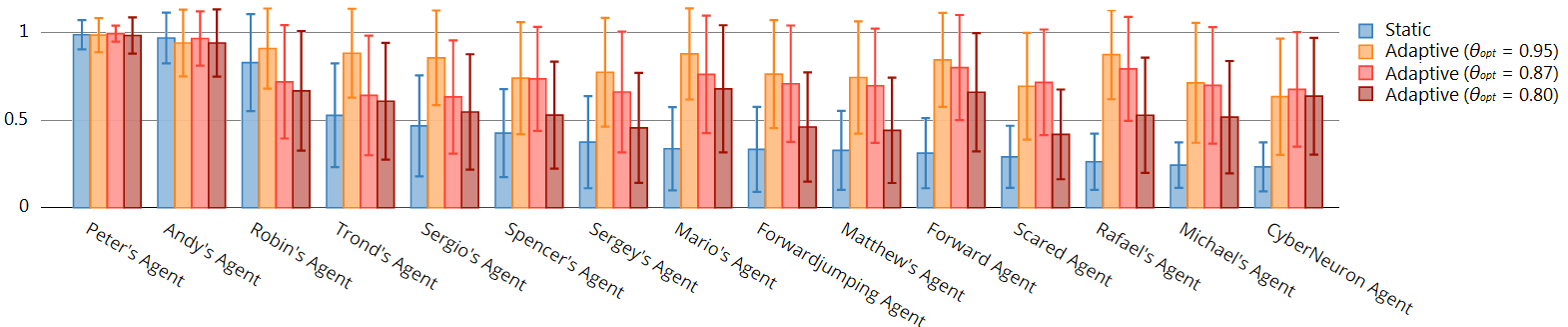}
\caption{Statistics (mean and standard deviation) of completion rates achieved by 15 agents on adaptive and static game sets.}
\label{fig:adpRes}
\end{figure*}

Fig.~\ref{fig:adpRes} illustrates the mean and the standard deviation of completion rates achieved on four game sets by 15 agents\footnote{The slight difference between the performance presented in \cite{TMAC} and that reported here is due to different Mario start state settings: the fire state in theirs but the small state in ours.}.
As shown in Fig.~\ref{fig:adpRes}, Peter's, Andy's and Robin's agents outperform other agents in terms of the averaging completion rate thanks to the A* algorithm used in their implementation. Hence, we regard these three agents as ``skilful players" and all the rest are ``novices".
It is observed from Fig.~\ref{fig:adpRes} that our real-time adaptation algorithm works well for all the novice agents given the fact that their completion rates on adaptive game sets are higher than that on the static game set, which implies easier levels were dynamically generated to improve their performance. In contrast, the DDA performance varies for three skillful agents. While the completion rates of Robin's agent are achieved as expected, our adaptation algorithm does not work well for Peter's and Andy's agents. According to \cite{TMAC}, these two agents can survive from nearly all the procedural levels no matter how difficult they are. Thus, our algorithm could hardly find games to challenge them. From Fig.~\ref{fig:adpRes}, it is also evident that a higher value of $\theta_{opt}$ generally leads to a higher completion rate, as required by DDA.

\begin{figure*}[htbp]
\small
\centering
\includegraphics[width=18cm]{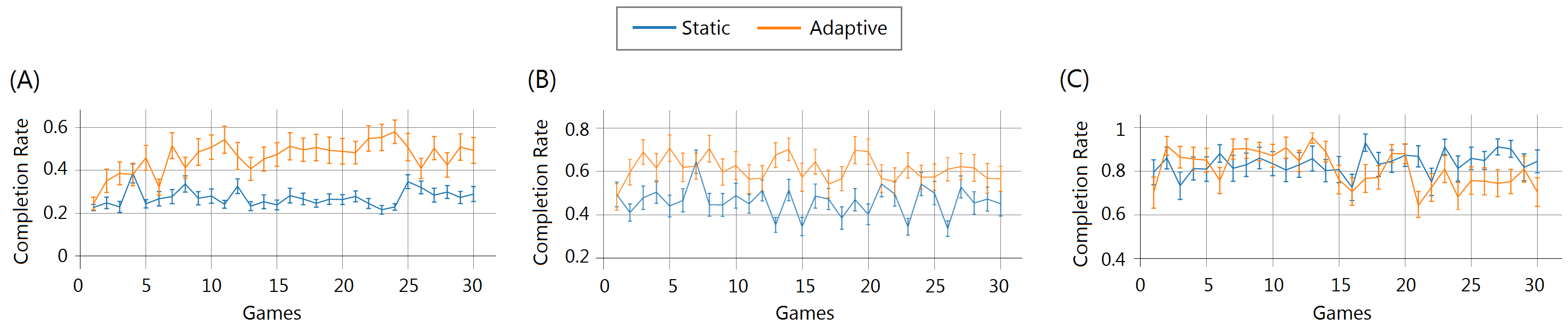}
\caption{Completion rates achieved by different agents in a gameplay episode of 30 games: adaptive vs. static. (A) Rafael's agent. (B) Sergio's agent. (C) Robin's agent.}
\label{fig:OLRes}
\end{figure*}

\begin{figure*}[htbp]
\small
\centering
\includegraphics[width=18cm]{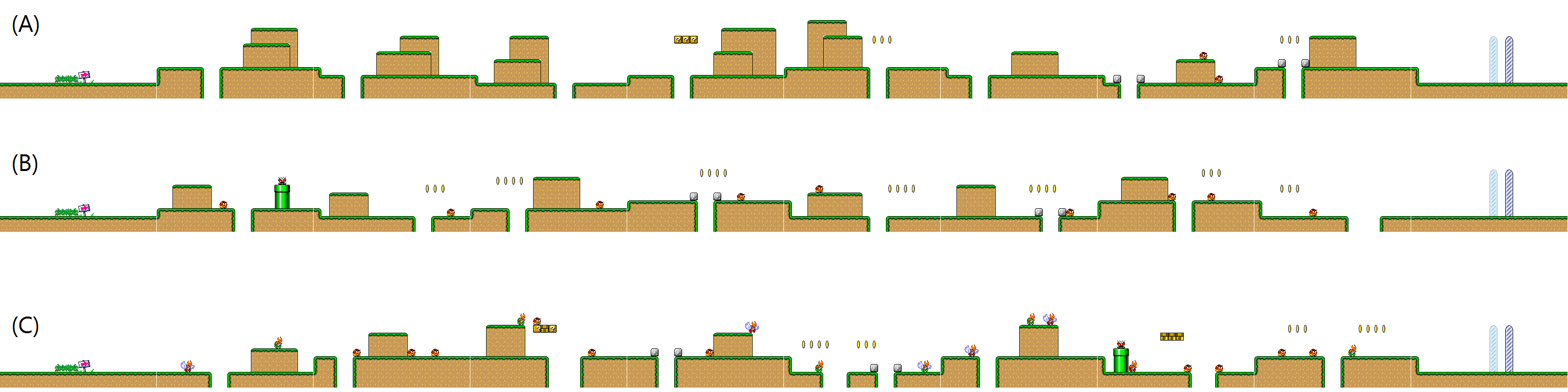}
\caption{Procedural level instances generated by our adaptation algorithm for different agents. (A) Rafael's agent. (B) Sergio's agent. (C) Robin's agent.}
\label{fig:PLA}
\end{figure*}

\par
To evaluate the online learning performance of our adaptation algorithm, we adopted
Rafael's, Sergio's and Robin's agents since they represent agents at different levels in light of playing styles and skills: Rafael's and Sergio's agents are novices and Robin's agent is  a skilful player.
In our experiment, each agent played 30 successive games generated in a gameplay episode by our adaptation algorithm via setting $\theta_{opt}=0.80$. For comparison, three agents also played 30 randomly generated static game levels of the same length. For reliability, we repeated this experiment for 30 trials and the mean and the standard derivation of completion rates achieved by three agents are illustrated in  Fig.~\ref{fig:OLRes}. From Fig.~\ref{fig:OLRes}(A) and (B), it is seen that the averaging completion rates of Rafael's and Sergio's agents on adaptive games are always higher than those on static games thanks to the adaptation that generates easier levels. In particular, the complete rates on adaptive games gradually increase and becomes stable after 5-8 games were played roughly. In contrast,  the completion rates of Robin's agent gradually decrease as observed from Fig.~\ref{fig:OLRes}(C) where the complete rates on adaptation games are always lower than those on their counterpart static games after 14 games were played thanks to the adaptation that keeps generate more difficult games.

\par
For demonstration, we exhibit three procedural levels generated by our adaptation algorithm for three aforementioned agents in Fig.~\ref{fig:PLA}. In the level for Rafael's agent shown in Fig.~\ref{fig:PLA}(A), there are fewer enemies than those generated for Sergio's and Robin's agents shown in Figs.~\ref{fig:PLA}(B) and~\ref{fig:PLA}(C). The level shown in Fig.~\ref{fig:PLA}(B) is at the intermediate difficult level where there are several enemies flower tubes, while the one shown in Fig.~\ref{fig:PLA}(C) contains a lot of enemies and gaps, a difficult level to play apparently.

\par
In summary, the experimental results reported in this section demonstrate that our proposed algorithms based on CPs work effectively in generating quality and adaptable SMB games.

\section{Discussion}

\par
In this section, we discuss the issues arising from our work and relate ours to pervious works.

\par
It is well known that automatically generated procedural SMB levels is generally worse than those handcrafted levels \cite{PCGGCAS}. This is further exacerbated by other simultaneous requirements in generation efficiency and controllability. By exploiting the nature of 2-D platform games, we effectively limit this problem to a smaller yet manageable content space consisting of short game segments without compromising the content diversity. Furthermore, we explore and exploit the synergy between rule-based and learning-based methodologies to produce quality building blocks, i.e., constructive primitives (CPs). While our hybrid quality assurance method appears effective, there is an issue to be addressed: how to trade-off between two methodologies. On the one hand, strict yet aggressive rules may completely remove all the unplayable content but often get rid of playable content as well so that the diversity of playable content is significantly limited.  Due to the high complexity of content space, it is extremely difficult to formulate rules, in particular, concerning all the aspects of game quality \cite{PCGG}. On the other hand, a learning-based approach may address all the quality assurance issues with a single evaluation function. It works efficiently with game developers' judgment on quality of training examples instead of handcrafting constraints and deductive rules based on the understanding of an entire content space \cite{LAUNCHPAD, AGACMPCVGL, ELSMBGE, AMLLG, PLGORE, MPESMB, NOTCH, FAMGCQ}. For example, the developer took only two hours in annotating 1900 games via visual inspection for our CP active learning as described in Sect. III. However, a learning-based approach rarely yields the error-free performance. Thus, our work presented in this paper suggests the ultimate goal for a hybrid approach as follows: without compromising the content diversity, efficient rule-based methods should concentrate on unplayability only to avoid catastrophic failures, and learning-based methods should take charge of other quality assurance aspects after unplayable content removal.
\par
While our online level generator clearly benefits from CPs in efficiency, it gains remarkable controllability via design elements, a direct content representation concerning low-level geometrical features, working at a local level for CPs. By controlling relevant design elements directly, ours generates a procedural level efficiently by integrating CPs of desired properties.
It is noticed that similar design elements were used as content representation in previous works \cite{AGACMPCVGL, ELSMBGE} where those attributes are specified at an entire procedural level.
As a result, such approaches have to formulate constructive rules or evaluation functions to control level properties globally. While those rules or evaluation functions have to be handcrafted with a great effort \cite{EDPCG, PCGG}, it may work less efficiently  especially for generating procedural levels of a considerable length \cite{ACEPLGM}. Thus, our approach significantly distinguishes those working at a global level \cite{AGACMPCVGL, ELSMBGE} in terms of content representation and resultant controllability.
\par
In general, our online level generator may be viewed as a hybrid PCG approach if we position it in light of the existing taxonomy \cite{SBPCG}. On the one hand, we use a generate-and-test method to produce CPs for quality assurance. On the other hand, a procedural level is constructively generated via a number of controllable parameters for efficiency \cite{PCGG}. Apparently, ours distinguishes from those  generate-and-test (e.g., \cite{LAUNCHPAD, AGACMPCVGL, ELSMBGE, AMLLG}) or constructive (e.g., \cite{EMGMC, MPESMB, NOTCH, FAMGCQ}) SMB level generators. As a hybrid approach, however, the desired level properties have to be specified via setting controllable parameters at a local level. This would be a potential weakness when such properties are unknown or hard to specify.
\par
For real-time DDA, our algorithm seems to resemble some previous works \cite{POLYMORPH, TCBPGS}. However, they differ in difficulty controllability and performance measurement. While a CP of certain difficulty is achieved by setting a controllable parameter in ours, a proper segment in theirs has to be generated via a rhythm-based mechanism or a set of constructive rules \cite{POLYMORPH, TCBPGS}. In addition, we use the survival rate, an objective metric, for measuring performance while they adopt the subjective player's feedback to decide the appropriateness of a difficulty level. While our CP-based adaptation algorithm is promising in performance-driven DDA, it is subject to a number of limitations: (a) the current evaluation is solely based on agents instead of human players and the adaptation process does not seem rapid; and (b) For DDA, the assumption used in our simulation, a strong relationship between player's performance and experience, may be questionable as suggested in recent studies (e.g., \cite{BCK}); (c) it is unclear how to adapt content via CPs in the presence of subjective feedback on an entire procedural level (e.g.,  \cite{MPESMB,TAPCGPG,EPCSMBGE}), which has been studied under the EDPCG framework \cite{EDPCG}.

\par
In conclusion, we have presented a novel approach to online level generation and real-time content adaptation in SMB via learning constructive primitives. Our approach has been evaluated via comparing to state-of-the-art SMB level generators in terms of quality, efficiency and controllability. Experimental results suggest that it meets the online level generation requirements and has a potential in real-time content adaptation. In our ongoing work, we are aiming addressing issues discussed above and further investigate our algorithms in different applications, e.g., experience-driven DDA. Although segment-based experience-driven content adaptation has been studied for SMB (e.g., \cite{TCBPGS,POLYMORPH}), their work entirely relies on the self-reported feedback on each short segment, which severely interrupts the gameplay experience \cite{EDPCG}. In our work, we are going to overcome this fundamental weakness by exploring player's behavior and relevant content features at the CP level for a scenario that only self-reported feedback on a complete procedural level is available. We anticipate that such real-time content adaptation minimizes gameplay interruption and yields optimal gameplay experience reciprocally.  Furthermore, we would like to investigate the feasibility in extending our approach proposed in this paper to other game genres such as first-person shooter.

\section*{Acknowledgment}

The authors are grateful to N. Shaker for providing SMB procedural level images used in our comparative study and to those members in the Google PCG Interest Group who responded to our enquiries for useful discussions.

\vbox{}

\ifCLASSOPTIONcaptionsoff
  \newpage
\fi

\end{document}